%% file: 00_main.tex
\pdfoutput=1

\documentclass[11pt]{article}

\usepackage{emnlp2021}

\usepackage{times}
\usepackage{latexsym}
\usepackage[T1]{fontenc}    
\usepackage[utf8]{inputenc} 
\usepackage{microtype}

\usepackage{hyperref}       
\usepackage{url}            
\usepackage{booktabs}       
\usepackage{amsfonts}       
\usepackage{nicefrac}       
\usepackage{xcolor}         
\usepackage{enumitem}
\usepackage{graphicx}
\usepackage{colortbl}
\usepackage{amsmath}
\usepackage{amssymb}
\usepackage{pifont}
\usepackage{multirow}
\usepackage{listings}
\definecolor{delim}{RGB}{20,105,176}
\definecolor{numb}{RGB}{106, 109, 32}
\definecolor{string}{rgb}{0.64,0.08,0.08}

\lstdefinelanguage{json}{
    numbers=right,
    numberstyle=\small,
    frame=single,
    rulecolor=\color{black},
    showspaces=false,
    showtabs=false,
    breaklines=true,
    postbreak=\raisebox{0ex}[0ex][0ex]{\ensuremath{\color{gray}\hookrightarrow\space}},
    breakatwhitespace=true,
    basicstyle=\ttfamily\small,
    upquote=true,
    morestring=[b]",
    stringstyle=\color{string},
    literate=
     *{0}{{{\color{numb}0}}}{1}
      {1}{{{\color{numb}1}}}{1}
      {2}{{{\color{numb}2}}}{1}
      {3}{{{\color{numb}3}}}{1}
      {4}{{{\color{numb}4}}}{1}
      {5}{{{\color{numb}5}}}{1}
      {6}{{{\color{numb}6}}}{1}
      {7}{{{\color{numb}7}}}{1}
      {8}{{{\color{numb}8}}}{1}
      {9}{{{\color{numb}9}}}{1}
      {\{}{{{\color{delim}{\{}}}}{1}
      {\}}{{{\color{delim}{\}}}}}{1}
      {[}{{{\color{delim}{[}}}}{1}
      {]}{{{\color{delim}{]}}}}{1},
}

\definecolor{purplish}{RGB}{99, 57, 116}
\newcommand{\textquote}[1]{\textit{#1}}
\newcommand{\emphcolor}[1]{\textcolor{teal}{#1}}
\newcommand{\splits}[1]{\textit{#1}}
\newcommand{\basemodels}[1]{\textsc{#1}}
\newcommand{\objclass}[1]{\texttt{#1}}
\newcommand{\action}[1]{\texttt{#1}}
\newcommand{\task}[1]{\textsc{#1}}

\newcommand{\modelfull}{Embodied BERT}
\newcommand{\modelac}{EmBERT}

\newcommand{\maskrcnn}{Mask RCNN}

\newcommand{\alfgoal}{\mathcal{I}_g}
\newcommand{\alfinst}{\mathcal{I}}
\newcommand{\alftraj}{\mathcal{T}}
\newcommand{\alfplan}{\mathcal{P}}
\newcommand{\alfaction}{a} 
\newcommand{\alfactionset}{\mathcal{A}} 
\newcommand{\alfobjmask}{M} 
\newcommand{\alffront}{\mathcal{V^F}} 
\newcommand{\alfleft}{\mathcal{V^L}} 
\newcommand{\alfback}{\mathcal{V^B}} 
\newcommand{\alfright}{\mathcal{V^R}} 

\newcommand{\cblkmark}{\ding{51}}
\newcommand{\cmark}{\raisebox{0pt}{\color{blue}{\ding{51}}}}
\newcommand{\xmark}{\raisebox{0pt}{\color{red}{\ding{55}}}}

\title{\modelfull: A Transformer Model for Embodied, Language-guided Visual Task Completion}

\author{%
  Alessandro Suglia$^1$
  \thanks{\hspace{3pt}Work completed via internship with Amazon Alexa AI.}
  \And
  Qiaozi Gao$^2$ 
  \And
  Jesse Thomason$^{2,3}$
  \AND
  Govind Thattai$^2$
  \And
  Gaurav S. Sukhatme$^{2,3}$ 
  \AND
  \textnormal{$^1$Heriot-Watt University; $^2$Amazon Alexa AI; $^3$University of Southern California}
}

\begin{document}

\maketitle

\begin{abstract}
Language-guided robots performing home and office tasks must navigate in and interact with the world.
Grounding language instructions against visual observations and actions to take in an environment is an open challenge.
We present \modelfull\ (\modelac), a transformer-based model which can attend to high-dimensional, multi-modal inputs across long temporal horizons for language-conditioned task completion.
Additionally, we bridge the gap between successful object-centric navigation models used for non-interactive agents and the language-guided visual task completion benchmark, ALFRED, by introducing object navigation targets for \modelac\ training.
\modelac\ achieves competitive performance on the ALFRED benchmark, and is the first model to use a full, pretrained BERT stack while handling the long-horizon, dense, multi-modal histories of ALFRED. Model code is available at the following link: \url{https://github.com/amazon-research/embert}
\end{abstract}

\section{Introduction}
\label{sec:introduction}
\input{01_introduction}

\section{Related Work}
\label{sec:related_work}
\input{02_related_work}

\section{The ALFRED Benchmark}
\label{sec:task}
\input{03_task}

\section{\modelfull}
\label{sec:model}
\input{04_model}

\section{Experiments and Results}
\label{sec:results}
\input{05_results}

\section{Conclusions}
\label{sec:conclusions}

\input{06_conclusions}


\section{Implications and Impact}
\label{sec:limitations}

\input{07_limitations}


\bibliographystyle{acl_natbib}
\bibliography{bibliography}



\clearpage

\appendix
\section{Appendix}
\label{sec:appendix}
\input{99_appendix}

\end{document}

%% file: 01_introduction.tex

Language is grounded in agent experience based on interactions with the world
~\cite{egl,bender-koller-2020-climbing}.
Task-oriented, instructional language focuses on \emph{objects} and \emph{interactions} between objects and actors, as seen in instructional datasets~\cite{Damen2020Collection,koupaee2018wikihow}, as a function of the inextricable relationship between language and objects~\cite{quine1960word}.
That focus yields language descriptions of object targets for manipulation such as \textquote{put the \emphcolor{strawberries} on the \emphcolor{cutting board} and slice	them into \emphcolor{pieces}}~\cite{chai:ijcai18}.
We demonstrate that predicting navigational object landmarks \emph{in addition} to manipulation object targets improves the performance of an instruction following agent in a rich, 3D simulated home environment.
We posit that object-centric navigation
is a key piece of semantic and topological navigation~\cite{kuipers1991robot} for Embodied AI (EAI) agents generally.

\begin{figure*}
\centering
\includegraphics[width=.9\linewidth]{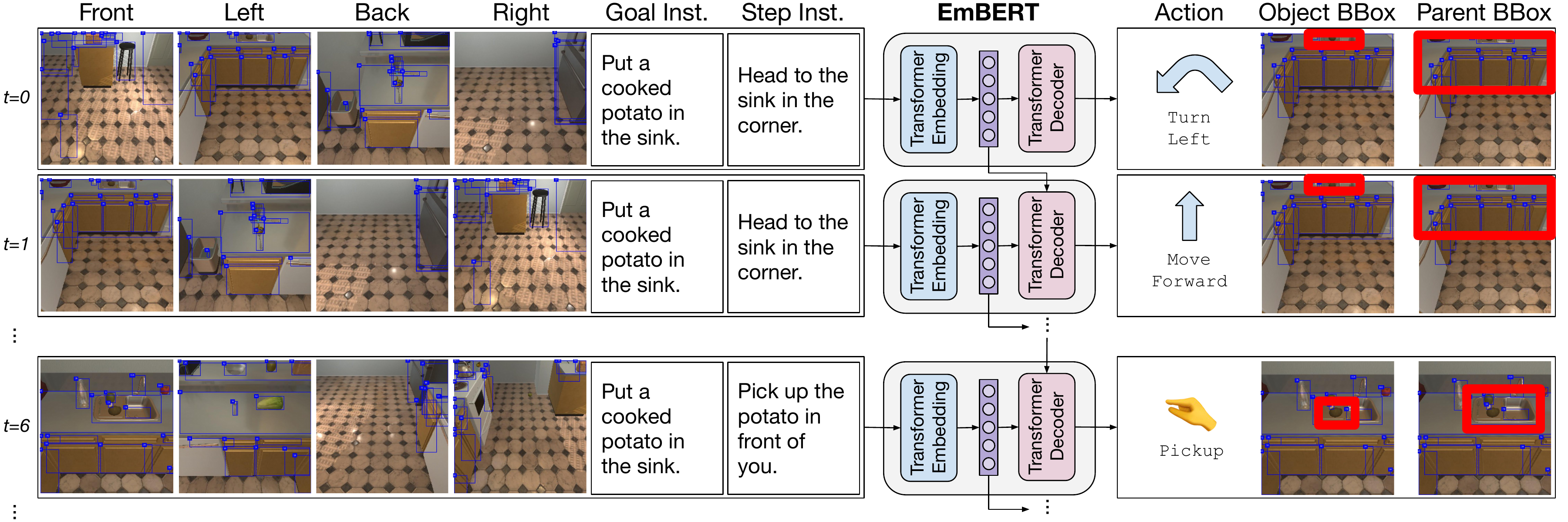}
\caption{\textbf{\modelfull.}
\modelac\ attends to object detections in a panoramic view around an agent, then predicts an action and both a target object and target object parent for both navigation and manipulation actions.
For example, at timesteps $t=0,1$ above, the model must predict the \emphcolor{\objclass{sink}} object target and its parent, the \emphcolor{\objclass{countertop}}, while at $t=6$ it predicts both the object \emphcolor{\objclass{potato}} to pick up and the \emphcolor{\objclass{sink}} on which it rests.
}
\label{fig:main}
\end{figure*}

Substantial modeling~\cite{majumdar:eccv20} and benchmark~\cite{reverie} efforts in EAI navigation focus on identifying object landmarks~\cite{blukis:corl18} and destinations~\cite{batra2020objectnav}.   
However, for agent \emph{task completion}, where agents must navigate an environment and manipulate objects towards a specified goal~\cite{iqa,alfred}, most 
predict movement actions without explicitly identifying \textit{navigation} object targets~\cite{moca,et,lwit,imitating_interactive_intelligence}.
We address this gap, grounding navigation instructions like \textquote{Head to the \emphcolor{sink} in the corner} by predicting the spatial locations of the goal \emphcolor{\objclass{sink}} object at each timestep (Figure~\ref{fig:main}).

Transformer-based models in EAI score the alignment between a language instruction and an already-completed path~\cite{majumdar:eccv20} or introduce recurrence by propagating part of the hidden state to the next timestep~\cite{cyclevlnbert}.
The former requires beam search over sequences of environment actions, which is not feasible when actions cannot be undone, such as \action{slicing} an \objclass{apple}.
The latter introduces a heavy memory requirement, and is feasible only with short trajectories of four to six steps.
We overcome both limitations by \emph{decoupling} the embedding of language and visual features from the prediction of what action to take next in the environment.
We first embed language and visual observations \emph{at single timesteps} using a multi-modal transformer architecture, then train a transformer decoder model to consume sequences of such embeddings to decode actions (Figure~\ref{fig:model}).

We introduce \modelfull\ (\modelac), which implements these two key insights:
\begin{enumerate}[noitemsep,nosep]
    \item \textbf{Object-centric Navigation} unifies the disjoint \emph{navigation} and \emph{interaction} action sequences in ALFRED, giving navigation actions per-step object landmarks.
    \item \textbf{Decoupled Multimodal Transformers} enable extending transformer based multimodal embeddings and sequence-to-sequence prediction to the fifty average steps present in ALFRED trajectories.
\end{enumerate}


%% file: 02_related_work.tex
Natural language guidance of robots~\cite{tellex:arcras:20} has been explored in contexts from furniture assembly~\cite{tellex2011}
to quadcoptor flight control~\cite{blukis:corl19}.

\textbf{Embodied AI.} 
For task completion benchmarks, actions like \action{pickup} must be coupled with object targets in the visual world, with specification ranging from mask prediction only~\cite{alfred} to proposals for full low level gripper control~\cite{rearrangement}.
Similarly, navigation benchmarks incorporate objects as targets in tasks like object navigation~\cite{reverie,batra2020objectnav,hiearchical_mechanical_search}, and explicitly modeling those objects assists generally at navigation success~\cite{visitron,Qi2020ObjectandActionAM,know-what-know-where}.
Many successful modeling approaches for navigation benchmarks incorporate multimodal transformer models that require large memory from recurrence~\cite{cyclevlnbert}, beam search over potential action sequences~\cite{majumdar:eccv20}, or shallow layers without large-scale pretraining to encode long histories~\cite{et,crossmap}.
In this work, we incorporate navigation object targets into the ALFRED task completion benchmark~\cite{alfred}, and decouple transformer-based multimodal state embedding from transformer-based translation of state embeddings to action and object target predictions. In addition, differently from other approaches that train from scratch their language encoder, we successfully exploit the BERT stack in our multi-modal architecture. In this way, \modelac{} can be applied to other language-guided tasks such as VLN and Cooperative Vision-and-Dialog Navigation~\cite{cvdn}. 

\input{tables/model_comparisons}

\textbf{Language-Guided Task Completion.} 
Table~\ref{tab:model_comparison} summarizes how \modelac\ compares to current ALFRED modeling approaches.
ALFRED language instructions are given as both a single high level goal and a sequence of step-by-step instructions (Figure~\ref{fig:data_aug}).
At each timestep, we encode the goal instruction and a predicted current step-by-step instruction.
We train \modelac\ to predict when to advance to the next instruction, a technique introduced by LWIT~\cite{lwit}.

\modelac\ uses a panoramic view space to see all around the agent.
Rather than processing dense, single vector representations~\cite{alfred,moca,et,abp,hlsm}, \modelac\ attends directly over object bounding box predictions embedded with their spatial relations to the agent, inspired by LWIT~\cite{lwit} and a recurrent VLN BERT model~\cite{cyclevlnbert}.
We similarly follow prior work~\cite{moca,et,lwit,abp,hitut} in predicting these bounding boxes as object targets for actions like \action{Pickup}, rather than directly predicting a dense object segmentation mask~\cite{alfred}.

Consider the step \textquote{heat the mug of water in the microwave}, where the visual observation before turning the \objclass{microwave} on and after turning the \objclass{microwave} off are identical.
Transformer encodings of ALFRED's large observation history are possible only with shallow networks~\cite{et} that cannot take advantage of large scale, pretrained language models used on shorter horizons~\cite{cyclevlnbert}.
We decouple multimodal transformer state encoding from sequence to sequence state to action prediction, drawing inspiration from the AllenNLP SQuAD~\cite{squad} training procedure~\cite{allennlp}. 

Our \modelac\ model is the first to utilize an auxiliary, object-centric navigation prediction loss during joint navigation and manipulation tasks, building on prior work that predicted only the \emph{direction} of the target object~\cite{shane-amazon} or honed in on landmarks during navigation-only tasks~\cite{visitron}.
While mapping environments during inference has shown promise on both VLN~\cite{scenememory,topoplan} and ALFRED~\cite{hlsm}, we leave the incorporation of mapping to future work.

%% file: tables/model_comparisons.tex
\begin{table*}
\centering
\begin{small}
\begin{tabular}{@{}lccccccccc@{}}
& \multicolumn{2}{c}{\bf Language Obs.} & \multicolumn{2}{c}{\bf Visual Obs.} & \multicolumn{2}{c}{\bf Historical Obs.} & \multicolumn{2}{c}{\bf Inference} \\
& Goal Inst. & Inst. & \multirow{2}{*}{Views} & \multirow{2}{*}{Features} & As & Hidden & Mask & Nav Obj. \\
& Structure & Split & & & Inputs & States & Pred. & Pred. \\
\toprule
\basemodels{Seq2Seq}~\cite{alfred} & \xmark & \xmark & Single & Dense & \xmark & LSTM & Direct & \xmark \\
\basemodels{MOCA}~\cite{moca} & \cmark & \xmark & Single & Dense & \xmark & LSTM & BBox & \xmark \\
\basemodels{ET}~\cite{et} & \xmark & \xmark & Single & Dense & TF & \xmark & BBox & \xmark \\
\basemodels{LWIT}~\cite{lwit} & \cmark & \cmark & Multi & BBox & \xmark & LSTM & BBox & \xmark \\
\basemodels{ABP}~\cite{abp} & \cmark & \xmark & Multi & Dense & \xmark & LSTM & BBox & \xmark \\
\basemodels{HiTUT}~\cite{hitut} & \cmark & \cmark & Single & BBox & SG & \xmark & BBox & \xmark \\
\basemodels{HLSM}~\cite{hlsm} & \cmark & - & Single & Dense & SG+Map & \xmark & Direct & \xmark \\
\midrule          
\modelac & \cmark & \cmark & Multi & BBox & \xmark & TF & BBox & \cmark \\
\bottomrule
\end{tabular}
\end{small}
\caption{\textbf{Model comparison.} \modelac\ uses a multimodal transformer (TF) to embed language \emph{inst}ructions and detected objects in a panoramic view, and a transformer decoder to produce action and object predictions.
Ours is the first ALFRED model to add object prediction to \textit{navigation} steps.
Other methods maintain history by taking previous transformer states (TF)~\cite{et}, subgoal prediction structures (SG)~\cite{hitut,hlsm}, or maintained voxel maps~\cite{hlsm} as input.
}
\label{tab:model_comparison}
\end{table*}

%% file: 03_task.tex
\begin{figure*}
\centering
\includegraphics[width=.9\linewidth]{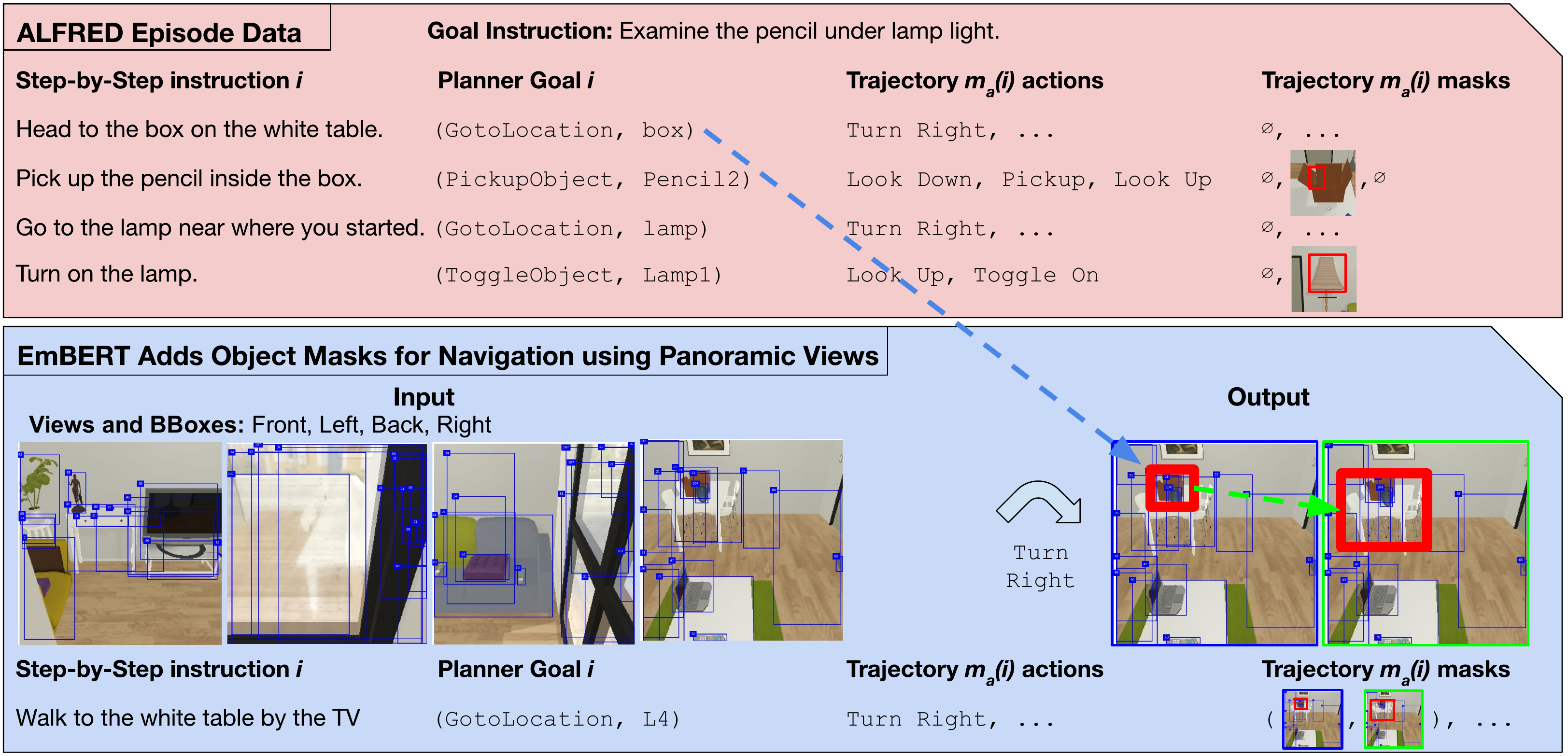}
\caption{\textbf{\modelac\ Auxiliary Predictions.}
ALFRED provides goal and step-by-step language instructions that are aligned with planner goals and sequences of trajectory actions in an expert demonstration (top).
\modelac\ additionally identifies \textit{navigational} object targets in a panoramic view (bottom).
\modelac\ predicts an object \textcolor{blue}{target} and its higher visibility parent \textcolor{green}{receptacle}, such as the \textcolor{green}{\objclass{table}} on which the \textcolor{blue}{\objclass{box}} rests.
}
\label{fig:data_aug}
\end{figure*}

The ALFRED benchmark~\cite{alfred} pairs household task demonstrations with written English instructions in 3d simulated rooms~\cite{ai2thor}.
ALFRED tasks are from seven categories: \task{pick \& place}, \task{stack \& place}, \task{pick two \& place}, \task{clean \& place}, \task{heat \& place}, \task{cool \& place}, and \task{examine in light}.
Each task involves one or more objects that need to be manipulated, for example an \objclass{apple}, and a final receptacle on which they should come to rest, for example a \objclass{plate}.
Many tasks involve intermediate state changes, for example \task{heat \& place} requires \emph{cooking} the target object in a \objclass{microwave}.

\textbf{Supervision Data.}
Each ALFRED episode comprises an initial state for a simulated room, language instructions, planning goals, and an expert demonstration trajectory.
The language instructions are given as a high-level goal instruction $\alfgoal$, for example \textquote{Put a cooked egg in the sink}, together with a sequence of step-by-step instructions $\vec{\alfinst}$, for example \textquote{Turn right and go to the sink}, \textquote{Pick up the egg on the counter to the right of the sink}, $\dots$
The planning goals $\alfplan$ (or \emph{sub-goals}) are tuples of goals and arguments, such as \texttt{(SliceObject, Apple)} that unpack to low-level sequences of actions like picking up a \objclass{knife}, performing a \action{slice} action on an \objclass{apple}, and putting the \objclass{knife} down on a \objclass{countertop}.
The expert demonstration trajectory $\alftraj$ is a sequence of action and object mask pairs, where $\alftraj_j=(\alfaction_j, \alfobjmask_j)$.
Each step-by-step instruction $\alfinst_i$ corresponds to a sub-sequence of the expert demonstration, $\alftraj_{j:k}$ given by alignment lookup $m_a(i)=(j,k)$ and to a planning goal $\alfplan_b$ by alignment lookup $m_p(i)=b$.
For example, in Figure~\ref{fig:data_aug}, instruction $\alfinst_0$ corresponds to a \texttt{GotoLocation} navigation goal, as well as a sequence of turning and movement API actions that a model must predict.

\textbf{Model Observations.}
At the beginning of each episode in timestep $t=0$, an ALFRED agent receives the high-level and step-by-step language instructions $\alfgoal,\vec{\alfinst}$.
At every timestep $t$, the agent receives a 2d, RGB visual observation representing the front-facing agent camera view, $\alffront$.
ALFRED models produce an action $a_t$ from among 5 navigation (e.g., \action{Turn Left}, \action{Move Forward}, \action{Look Up}) and 7 manipulation actions (e.g., \action{Pickup}, \action{ToggleOn}, \action{Slice}), as well as an object mask $M_t$.
Predicted action $a_t$ and mask $M_t$ are executed in the ALFRED environment to yield the next visual observation.
For navigation actions, prediction $M_t$ is ignored, and there is no training supervision for objects associated with navigation actions.

\textbf{\modelac\ Predictions.} \label{par:data_aug}
\modelac\ gathers additional visual data (Figure~\ref{fig:data_aug}). 
After every navigation action, we turn the agent in place to obtain left, backwards, and right visual frames $\alfleft$, $\alfback$, $\alfright$.
Following prior work~\cite{moca}, we run a pretrained Mask-RCNN~\cite{maskrcnn} model to extract bounding boxes from our visual observations at each view.
We train \modelac\ to select the bounding box which has the highest intersection-over-union with $M_t$ (more details in Section~\ref{sec:model}).

We define a \emph{navigation object target} for navigation actions.
For navigation actions taken during language instruction $\alfinst_i$, we examine the frame $\alffront_k$ at time $k$ for $\alftraj_k$; $m_a(i)=(j,k)$.
We identify the object instance $O$ of the class specified in the planning goal $\alfplan_{m_b(i)}$ in $\alffront_k$.
We define this object $O$ as the navigation object target for all navigation actions in $\alftraj_{j:k}$ by pairing those actions with object mask $\alfobjmask^O$ to be predicted during training.
We also add a training objective to predict the \emph{parent receptacle} $P(O)$ of $O$.
Parent prediction enables navigating to landmarks such as the \objclass{table} for instructions like \textquote{Turn around and head to the box on the table}, where the \objclass{box} compared to the table on which it rests (Figure~\ref{fig:data_aug}).

%% file: 04_model.tex
\begin{figure*}
\centering
\includegraphics[width=.9\linewidth]{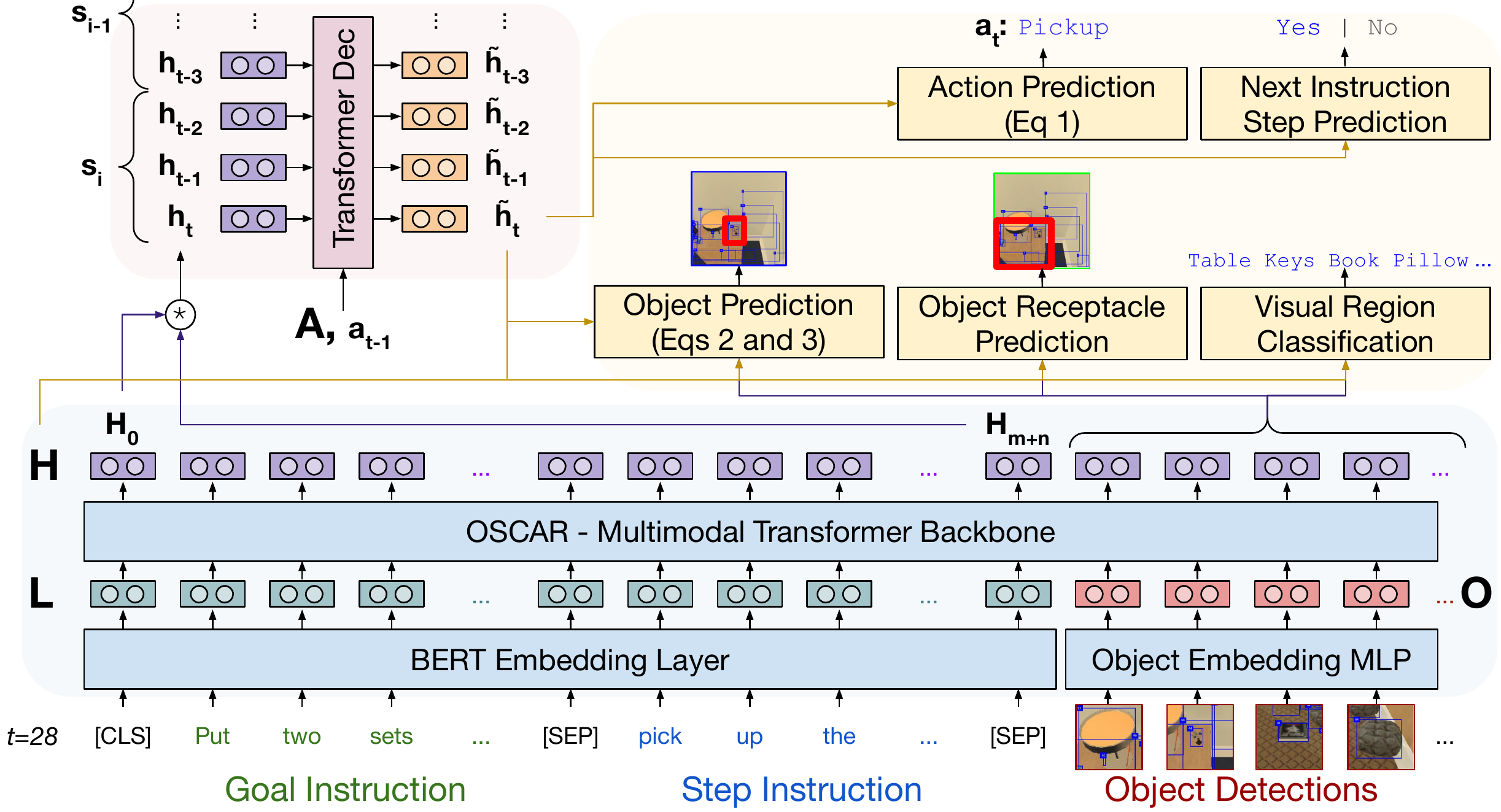}
\caption{\textbf{Proposed \modelfull\ model.}
A \textcolor{blue}{multimodal encoder} embeds goal- and step-level instructions alongside object detections from a panoramic view around the agent.
This encoder produces a temporally independent hidden state $h_t$.
A sequence of such hidden states are attended by a \textcolor{red}{segment-level recurrent action decoder} to produce time-dependent states $\tilde{h}_t$.
\modelac\ is trained in segments $s_i$ to balance gradient flow over time with memory constraints, and previous segments are cached to be attended over in future timesteps.
Time-dependent state $\tilde{h}_t$ is used to \textcolor{orange}{predict} the next action, whether to start attending to the next step-by-step instruction, what object to target in the environment, that object's parent receptacle, and detected object classes.
}
\label{fig:model}
\end{figure*}

\modelac\ uses a transformer encoder for jointly embedding language and visual tokens and an transformer decoder for long-horizon planning and object-centric navigation predictions (Figure~\ref{fig:model}).

\subsection{Multimodal encoder}
\label{ssec:mm_encoder}

We use OSCAR~\cite{oscar} as a backbone transformer module to fuse language and visual features at each ALFRED trajectory step. 
We obtain subword tokens for the goal instruction $\alfgoal = \{g_1, g_2, \dots, g_n\}$ and the step-by-step instruction $\alfinst_j = \{i_1, i_2, \dots, i_m\}$ using the WordPiece tokenizer~\cite{wu2016google} and process the sequence as: \texttt{[CLS]} $\alfgoal$ \texttt{[SEP]} $\alfinst_j$ \texttt{[SEP]}, using token type ids to distinguish the goal and step instructions.
We derive token embeddings $\mathbf{L} \in \mathcal{R}^{(m+n+3) \times d_e}$ using the BERT~\cite{bert} embedding layer, where $d_e$ is the embedding dimensionality. 

We provide EmBERT with object-centric representations by using MaskRCNN~\cite{maskrcnn} features to represent detected objects in every frame of the panorama view.
We freeze the weights of a MaskRCNN model fine-tuned for AI2-THOR frames~\cite{moca}.
We fix the number of object detections in the front view $\alffront$ to $36$, while limiting those in the side views to $18$. 
We represent each object $o \in \mathcal{O}$ as an embedding $\mathbf{o} \in \mathbb{R}^{d_o}$, which is a concatenation of: 1) detection ResNet~\cite{resnet} features; 2) bounding box coordinates; 3) bounding box relative area; and 4) vertical and horizontal heading of the object related to the current agent position, following prior work~\cite{shane-amazon}.
These representations make up the observed object embeddings $\mathbf{O}$. 
We use a one layer MLP to map object embeddings of dimensionality $d_o$ to size $d_e$.\footnote{In our experiments, in order to reuse the visual embedding available in the OSCAR checkpoint, we use an additional one layer MLP to adapt our visual features to the visual embeddings space learned by OSCAR.}
The multi-modal transformer backbone consumes the token and object embeddings to produce multi-modal hidden states $\mathbf{H} \in \mathbb{R}^{m+n+|O| \times d_e}$. 
We obtain these state representations, $\mathbf{h}_t$, for each timestep $t$ by computing an element-wise product between $\mathbf{H}_0$ and $\mathbf{H}_{m+n}$, the hidden state of the \texttt{[CLS]} token and the last \texttt{[SEP]} token placed between language tokens and objects, similar in spirit to the approach described in \cite{zhou2020unified}.
In this way, we can generate \emph{temporally independent} agent states for an entire trajectory resulting in a sequence of states $\{\mathbf{h}_1, \mathbf{h}_2, \dots, \mathbf{h}_{|\alftraj|}\}$.

\subsection{Segment-Level Recurrent Action Decoder}
The ALFRED challenge requires models to learn to complete action sequences averaging 50 steps and spanning multiple navigation and manipulation sub-goals.
However, due to the quadratic complexity of the self-attention mechanism, feeding long sequences to transformers is computationally expensive~\cite{beltagy2020longformer}.
Inspired by the TransformerXL model~\cite{dai2019transformer}, we design the Segment-Level Recurrent Action Decoder architecture that models long trajectories with recurrent segment-level state reuse.
At training time we divide trajectories into temporal segments of size $s$. 
Given two consecutive segments, $\mathbf{s}_i$ and $\mathbf{s}_{i+1}$, \modelac\ caches the representations generated for segment $\mathbf{s}_i$.
The computed gradient does not flow from $\mathbf{s}_{i+1}$ to $\mathbf{s}_i$, but cached representations are used as extended context. 
When predicting the next action, the model can still perform self-attention over the previous segment representations, effectively incorporating additional contextual information that spans an high number of previous timesteps.

The TransformerXL model is intended as an encoder-only architecture which is not able to perform cross-attention with some encoder hidden states. 
Therefore, we introduce two novel elements to its architecture: 1) \textit{encoder hidden states cache}; 2) \textit{cross-attention over encoder states}.
First, our extended context is composed of both agent state representations and hidden states from the previous segment $\mathbf{s}_i$.
In addition, to perform cross-attention between decoder and encoder hidden states, we modify the TransformerXL self-attention mechanism following common practice in designing transformer decoders~\cite{vaswani2017attention}. 
\modelac\ encodes the previous actions for the current timestep $a_{t-1}$ and extracts an action embedding $\mathbf{a}_t$ from a learnable embedding matrix $\mathbf{A} \in \mathbb{R}^{|\alfactionset| \times d_a}$.
In the TransformerXL's multi-head self-attention layers, we generate \emph{keys} and \emph{values} from the agent state representations (\emph{encoder}) and \emph{queries} from the action embeddings (\emph{decoder}).
We obtain \textit{time-dependent} agent state representations $\{\mathbf{\tilde{h}}_1, \mathbf{\tilde{h}}_2, \dots, \mathbf{\tilde{h}}_{|\alftraj|}\}$ as output.

Given time-dependent hidden states, the model predicts action and object mask outputs. 
We learn a probability distribution over the agent actions $\alfactionset$ by using a two layer feedforward network (FFN) with dropout and GeLU~\cite{hendrycks2016gaussian} activation receiving the hidden state $\mathbf{\tilde{h}}_t$ for the timestep $t$:
\begingroup\makeatletter\def\f@size{8}\check@mathfonts
\def\maketag@@@#1{\hbox{\m@th\large\normalfont#1}}%
\begin{align}\label{eq:action_ffn}
    \mathbf{\tilde{h}}^1_t &= \text{GeLU}(\mathbf{\tilde{h}}_t \mathbf{W}^1) &
    P(a_t | \mathbf{\tilde{h}}_t) &=  \text{softmax}(\mathbf{\tilde{h}}^1_t \mathbf{W}^2),
\end{align}
\endgroup
where $\mathbf{W}^1 \in \mathbb{R}^{d_e \times d_e}$ and $\mathbf{W}^2 \in \mathbb{R}^{d_e \times |\alfactionset|}$ are two weight matrices. 
We use sequence-based cross-entropy loss~\cite{sutskever2014sequence}, $\mathcal{L}_A$, to supervise the action prediction task.
In addition, we derive \emph{time-dependent} fine-grained representations of token and object embeddings. We use conditional scaling~\cite{dumoulin2018feature} to fuse the decoder hidden state $\mathbf{\tilde{h}}_t$ with the embedding $\mathbf{H}$ to produce the time-dependent embeddings $\mathbf{\tilde{H}}$:
\begingroup\makeatletter\def\f@size{8}\check@mathfonts
\def\maketag@@@#1{\hbox{\m@th\large\normalfont#1}}%
\begin{align}\label{eq:conditional_scaling}
    \mathbf{\tilde{c}} &= \mathbf{W}_t \mathbf{\tilde{h}} &
    \mathbf{\tilde{H}}_i &= \mathbf{\tilde{c}} \cdot \mathbf{H}_i,_{i = \{1, \dots, (m+n+|O|)\}},
\end{align}
\endgroup
where $\mathbf{W}_t \in \mathbb{R}^{d_e \times d_e}$ is a weight matrix used to adapt the representation of the original decoder hidden state $\mathbf{\tilde{h}}$. 
We predict target objects by selecting one bounding box among the detections in $\alffront$ for manipulation actions, or any view for navigation actions.
We treat object mask prediction as a classification task where the model first extracts time-dependent object embeddings $\mathbf{\tilde{O}} = \mathbf{\tilde{H}}_i$, $i = \{(m+n), \dots, (m+n+|O|)\}$, and then generates logits for each object as follows:
\begingroup\makeatletter\def\f@size{8}\check@mathfonts
\def\maketag@@@#1{\hbox{\m@th\large\normalfont#1}}%
\begin{align}\label{eq:object_logits}
    \mathbf{\tilde{o}}^1_i &= \text{GeLU}(\mathbf{\tilde{O}} \mathbf{W}^1_o) &
    P(o_i | \mathbf{\tilde{O}}_i) &=  \text{softmax}(\mathbf{\tilde{o}}^1_i \mathbf{W}^2_o),
\end{align}
\endgroup
where $\mathbf{W}^1_o \in \mathbb{R}^{d_e \times d_e}$ and $\mathbf{W}^2_o \in \mathbb{R}^{d_e \times 1}$ are two weight matrices. 
At training time, we determine the target object by using the Intersection-Over-Union score between the predicted object masks generated by MaskRCNN for each object and the gold object mask.
To supervise this classification task, we use sequence-based cross-entropy loss, $\mathcal{L}_O$.

\subsection{Auxiliary tasks}

During the \modelac\ training, we jointly optimize $\mathcal{L}_A$, $\mathcal{L}_O$, and several auxiliary tasks.

\textbf{Next Instruction Prediction.}
Several existing models for ALFRED encode the sequence of language instructions $\alfinst$ together with the goal (Table~\ref{tab:model_comparison}), or concatenate step-by-step instructions. 
These simplifications can prevent the model from carefully attending to relevant parts of the visual scene.
\modelac\ takes the first instruction at time $t=0$, and performs add an auxiliary prediction task to advance from instruction $\alfinst_{j}$ to instruction $\alfinst_{j+1}$.
To supervise the next-instruction decision, we create a binary label for each step of the trajectory that indicates whether that step is the last step for a specific sub-goal, as obtained by $m_a(i)$.
We use a similar FNN as Equation~\ref{eq:action_ffn}to model a Bernoulli variable used to decide when to advance to the next instruction.
We denote the binary cross-entropy loss used to supervise this task as $\mathcal{L}_{INST}$. 

\textbf{Object Target Predictions.}
\modelac\ predicts a target object for navigation actions, 
together with the receptacle object containing the target, for example a \objclass{table} on which a \objclass{box} sits (Figure~\ref{fig:data_aug}).
For these tasks, we use an equivalent prediction layer to the one used for object prediction.
We denote the cross-entropy loss associated with these task by $\mathcal{L}_{NAV}$ and $\mathcal{L}_{RECP}$. 

\textbf{Visual Region Classification.}
Class-conditioned representations are useful for agent manipulation, especially when combined with hand-crafted procedures for object selections~\cite{moca}.
Inspired by masked region modeling tasks~\cite{uniter,visitron}, we select with $\%15$ probability some objects part of the agent view in a given timestep $t$ and we ask the model to predict their classes.
Given the instruction \textquote{Turn around and walk to the book on the desk}, at the very first timestep of the trajectory it is likely that none of the mentioned objects are visible.
Thus, we assume that at the last step of a sub-goal the agent will have in view the objects associated with the instruction. For the prediction task, we directly use the time-dependent object embeddings $\mathbf{\tilde{O}}$ and use an FFN (similar to Equation~\ref{eq:action_ffn}) to estimate a probability distribution over the ALFRED object labels. We use a cross-entropy loss denoted by $\mathcal{L}_{VRC}$ as supervision for this task. 

%% file: 05_results.tex
\input{tables/main_results}

\modelac\ achieves competitive performance with state of the art models on the ALFRED leaderboard test sets (Table~\ref{tab:results}), surpassing all but ET~\cite{et} and ABP~\cite{abp} on \splits{Seen} test fold performance (Table~\ref{tab:full_ablations}) at the time of writing.
Notably, \modelac\ achieves this performance without augmenting ALFRED data with additional language instructions, as is done in ET~\cite{et}, or visual distortion as used in ABP~\cite{abp}.

\input{tables/full_ablations}

\textbf{Implementation Details.} \label{par:impl_details}
\modelac\ is implemented using AllenNLP~\cite{allennlp}, PyTorch-Lightning,\footnote{\url{https://www.pytorchlightning.ai/}} and Huggingface-Transformers~\cite{wolf2019huggingface}. 
We train using the Adam optimizer with weight fix~\cite{loshchilov2017decoupled}, learning rate $2e^{-5}$, and linear rate scheduler without warmup steps. 
We use dropout of $0.1$ for the hidden layers of the FFN modules and gradient clipping of $1.0$ for the overall model weights. 
Our TransformerXL-based decoder is composed of $2$ layers, $8$ attention heads, and uses a memory cache of $200$ slots. 
At training time, we segment the trajectory into $10$ timesteps. 
In order to optimize memory consumption, we use bucketing based on the trajectory length. 
We use teacher forcing~\cite{williams1989learning} to supervise \modelac\ during the training process. 
To decide when to stop training, we monitor the average between action and object selection accuracy for every timestep based on gold trajectories. 
The best epoch according to that metric computed on the \emph{validation seen} set is used for evaluation.
The total time for each epoch is about 1 hour for a total of 20 hours for each model configuration using EC2 instances \texttt{p3.8xlarge} using 1 GPU.

\textbf{Action Recovery Module.}
For obstacle avoidance, if a navigation action fails, for example the agent choosing \action{MoveAhead} when facing a wall, we take the next most confident navigation action at the following timestep, as in MOCA~\cite{moca}.
We introduce an analogous object interaction recovery procedure.
When the agent chooses an interaction action such as \action{Slice}, we first select the bounding box of highest confidence to retrieve an object interaction mask.
If the resulting API action fails, for example if the agent attempts to \action{Slice} a \action{Kettle} object, we choose the next highest confidence bounding box at the following timestep. 
The ALFRED challenge ends an episode when an agent causes 10 such API action failures.

\textbf{Comparison to Other Models.}
Table~\ref{tab:results} gives \modelac\ performance against top and baseline models on the ALFRED leaderboard at the time of writing.
\splits{Seen} and \splits{Unseen} sets refer to tasks in rooms that were or were not seen by the agent at training time.
We report \textit{Task} success rate and \textit{Goal Conditioned (GC)} success rate.
Task success rate is the average number of episodes completed successfully.
Goal conditioned success rate is more forgiving; each episode is scored in $[0, 1]$ based on the number of subgoals satisfied, for example, in a \task{stack \& place} task if one of two \objclass{mugs} are put on a \objclass{table}, the GC score is $0.5$~\cite{alfred}.
Path weighted success penalizes taking more than the number of expert actions necessary for the task.

\modelac\ outperforms MOCA~\cite{moca} on \splits{Unseen} scenes, and several models on \splits{Seen} scenes.
The primary leaderboard metric is \splits{Unseen} success rate, measuring models' generalization abilities.
Among competitive models, \modelac\ outperforms only MOCA at \splits{Unseen} generalization success.
Notably, \modelac\ remains competitive on \splits{Unseen} \textit{path-weighted} metrics, because it does not perform any kind of exploration or mapping as in HLSM~\cite{hlsm} and ABP~\cite{abp}.

We do not utilize the MOCA \emph{Instance Association in Time} module~\cite{moca} that is mimicked by ET~\cite{et}.
That module is conditioned based on the object class of the target object selected across timesteps.
Because we directly predict object \textit{instances} without conditioning on a predicted object class, our model must learn instance associations temporally in an implicit manner, rather than using such an inference time ``fix''.

\textbf{\modelac\ Ablations.}
Removing the object-centric navigation prediction unique to \modelac\ decreases performance on all metrics (Table~\ref{tab:full_ablations}).
We show that limiting memory for the action decoder to a single previous timestep, initializing with BERT rather than OSCAR weights, and limiting vision to the front view all decrease performance in both \splits{Seen} and \splits{Unseen} folds.

We find that our parent prediction and visual region classification losses, however, do not improve performance.
To investigate whether a smaller model would benefit more from these two auxiliary losses, we ran \modelac\ with only $9$ bounding boxes per side view, which enables fitting longer training segments in memory (we use $14$ timesteps, rather than $10$).
We found that those losses improved \modelac\ performance on the \splits{Unseen} environment via both success rate and goal conditions metrics, and improved success rate alone in \splits{Seen} environments when the non-frontal views were limited to 9, rather than 18, bounding boxes.
Given the similar performance of \modelac\ with all three auxiliary losses at 18 and 9 side views, we believe \modelac\ is over-parameterized with the additional losses and 18 side view bounding boxes.
It is possible that data augmentation efforts to increase the volume of ALFRED training data, such as those in ET~\cite{et}, would enable us to take advantage of the larger \modelac\ configuration.

%% file: tables/main_results.tex
\newcommand{\mcc}[2]{\multicolumn{#1}{c}{#2}}
\newcommand{\mcp}[2]{\multicolumn{#1}{c@{\hspace{30pt}}}{#2}}
\definecolor{Gray}{gray}{0.90}
\newcolumntype{a}{>{\columncolor{Gray}}r}
\newcolumntype{g}{>{\columncolor{Gray}}c}
\newcommand{\B}[1]{\textcolor{blue}{\textbf{#1}}}

 \begin{table*}
    \setlength{\aboverulesep}{0pt}
    \setlength{\belowrulesep}{0pt}
    \centering
    \begin{small}
    \begin{tabular}{@{}laarr@{}}
                     & \mcc{4}{\textbf{Leaderboard Test Fold Performance}}                       \\
                     & \mcc{2}{\splits{Seen}}   & \mcc{2}{\splits{Unseen}}  \\
    Model            & \multicolumn{1}{g}{Task (PLW)} & \multicolumn{1}{g}{GC (PLW)} 
                     & \multicolumn{1}{c}{Task (PLW)} & \multicolumn{1}{c}{GC (PLW)} \\
    \toprule
    \addlinespace[-\aboverulesep]     
    \basemodels{Seq2Seq}~\cite{alfred} & \phantom{0}3.98 (\phantom{0}2.02)    & \phantom{0}9.42 (\phantom{0}6.27)    & \phantom{00}.39 (\phantom{0}0.08)   & \phantom{0}7.03 (\phantom{0}4.26) \\[1pt]
    \basemodels{HiTUT}~\cite{hitut} & 21.27 (11.10) & 29.97 (17.41) & 13.87 (\textbf{\phantom{0}5.86}) & 20.31 (11.51) \\[1pt]
    \basemodels{MOCA}~\cite{moca}        & 22.05 (15.10)    & 28.29  (22.05)    & \phantom{0}5.30 (\phantom{0}2.72)   & 14.28 (\phantom{0}9.99) \\[1pt]
    \basemodels{HLSM}~\cite{hlsm} & 25.11 (\phantom{0}6.69) & 35.79 (11.53) & \textbf{16.29} (\phantom{0}4.34) & \textbf{27.24} (\phantom{0}8.45) \\[1pt]
    \basemodels{LWIT}~\cite{lwit} & 30.92 (25.90)   & 40.53 (\textbf{36.76})   & \phantom{0}9.42 (\phantom{0}5.60) & 20.91 (\textbf{16.34}) \\[1pt]
    \bf \basemodels{\modelac}                    &  31.77 (23.41)   &  39.27 (31.32)    & \phantom{0}7.52 (\phantom{0}3.58)   & 16.33 (10.42) \\[1pt]
    \basemodels{ET}~\cite{et}      & 38.42 (\textbf{27.78})    & 45.44 (34.93)    & \phantom{0}8.57 (\phantom{0}4.10)   & 18.56 (11.46) \\[1pt]    
    \basemodels{ABP}~\cite{abp} & \textbf{44.55} (\phantom{0}3.88) & \textbf{51.13} (\phantom{0}4.92) & 15.43 (\phantom{0}1.08) & 24.76 (\phantom{0}2.22) \\[1pt]
    \addlinespace[-\belowrulesep] 
    \bottomrule
    \end{tabular}
    \end{small}
    \caption{\textbf{Test Fold Performance.}
    Path weighted metrics are given in parentheses.
    }
    \label{tab:results}
\end{table*}

%% file: tables/full_ablations.tex
\begin{table*}
    \small
    \setlength{\aboverulesep}{0pt}
    \setlength{\belowrulesep}{0pt}
    \centering
    \begin{tabular}{@{}c@{\hskip 4pt}c@{\hskip 4pt}c@{\hskip 4pt}c@{\hskip 4pt}c@{\hskip 4pt}crrrr@{}}
                     & & & & & & \mcp{4}{\textbf{Validation Fold Performance}} \\
                     \mcc{6}{\basemodels{\modelac}} & \mcc{2}{\splits{Seen}}   & \mcc{2}{\splits{Unseen}}  \\
                     Init Weights & \#SB & Mem & Nav $O$ & $P(O)$ & VRC & \multicolumn{1}{c}{Task} & \multicolumn{1}{c}{GC} &
                     \multicolumn{1}{c}{Task} & \multicolumn{1}{c}{GC} \\
    \toprule
    \addlinespace[-\aboverulesep]
    \midrule
    OSCAR & 18 & 200 & \cblkmark & \cblkmark & \cblkmark & 28.54 (22.88) & 38.69 (31.28) & \phantom{0}1.46 (\phantom{00}.72) & 10.19 (\phantom{0}6.25) \\[1pt]
    OSCAR & 18 & 200 & \cblkmark & & \cblkmark & 34.76 (28.46) & 41.30 (35.50) & \phantom{0}3.66 (\phantom{0}1.55) & 12.61 (\phantom{0}7.49) \\[1pt]
    OSCAR & 18 & 200 & \cblkmark & \cblkmark & & 36.22 (27.05) & 44.57 (35.23) & \phantom{0}4.39 (\phantom{0}2.21) & 13.03 (\phantom{0}7.54) \\[1pt]
    OSCAR & 18 & 200 & \cblkmark & & & \B{37.44} (\B{28.81}) & \B{44.62} (\B{36.41}) & \B{\phantom{0}5.73} (\B{\phantom{0}3.09}) & \B{15.91} (\B{\phantom{0}9.33}) \\[1pt]
    OSCAR & 18 & 200 & & & & 23.66 (17.62) & 29.97 (24.16) & \phantom{0}2.31 (\phantom{0}1.24) & 12.08 (\phantom{0}7.62) \\[1pt]
    BERT & 18 & 200 & \cblkmark & \cblkmark & &  26.46 (19.41)   &  35.70 (27.04)    & \phantom{0}3.53 (\phantom{0}1.77)   & 13.02 (\phantom{0}7.57)  \\[1pt]
    \midrule
    OSCAR & 9 & 200 & \cblkmark & \cblkmark & \cblkmark & 29.30 (20.14)   &  36.28 (27.21)    & \phantom{0}3.06 (\phantom{0}1.13)   & 12.17 (\phantom{0}6.69) \\[1pt]
    OSCAR & 9 & 200 & \cblkmark & \cblkmark & & 31.75 (23.52)   &  38.80 (32.21)    & \phantom{0}2.56 (\phantom{0}1.28)   & 12.97 (\phantom{0}8.24)  \\[1pt]
    OSCAR & 9 & 200 & & \cblkmark & \cblkmark & 20.37 (16.30)   &  28.64 (23.11)    & 1.46 (\phantom{0}0.75)   & 10.47 (\phantom{0}6.26)  \\[1pt]
    OSCAR & 9 & 200 & \cblkmark & &  & 28.33 (20.77)   &  36.83 (28.03)    & \phantom{0}2.68 (\phantom{0}1.18)   & 11.60 (\phantom{0}6.78)  \\[1pt]
    OSCAR & 9 & 200 & & & & 27.84 (20.66)   &  36.59 (27.97)    & \phantom{0}2.44 (\phantom{0}1.06)   & 11.46 (\phantom{0}6.76)  \\[1pt]
    \midrule
    OSCAR & 0 & 200 & \cblkmark & \cblkmark & &  25.31 (18.79)   &  34.27 (26.09)    & \phantom{0}3.42 (\phantom{0}1.49)   & 12.25 (\phantom{0}7.34)  \\[1pt]
    OSCAR & 9 & 1 & \cblkmark & \cblkmark & & 20.98 (13.98)   &  33.33 (22.74)    & \phantom{0}1.10 (\phantom{0}0.60)   & 10.33 (\phantom{0}4.69)  \\[1pt]
    OSCAR & 18 & 1 & \cblkmark & \cblkmark & & 21.95 (12.99) & 35.04 (22.31) & \phantom{0}1.58 (\phantom{00}.54) & 11.08 (\phantom{0}6.18) \\[1pt]
    \addlinespace[-\belowrulesep] 
    \bottomrule
    \mcc{6}{\basemodels{MOCA}~\cite{moca}} & 18.90 (13.20)   &  28.02 (21.81)    & \phantom{0}3.65 (\phantom{0}1.94)   & 13.63 (\phantom{0}8.50) \\[1pt]
    \bottomrule
    \end{tabular}
    \caption{\textbf{Validation Fold Performance.}
    We present ablations adjusting the number of side-view bounding boxes, attended memory length, with and without predicting navigation target $O$, target parent object $P(O)$, and visual region classification (VRC) loss. 
    We also explore initializing our multi-modal encoder with BERT versus OSCAR initialization.
    The highest values per fold and metric are shown in \B{blue}. Path weighted metrics are given in parenthesis.
    }
    \label{tab:full_ablations}
\end{table*}

%% file: 06_conclusions.tex
We apply the insight that object-centric navigation is helpful for language-guided Embodied AI to a benchmark of tasks in home environments.
Our proposed \modelfull\ (\modelac) model adapts the pretrained language model transformer OSCAR~\cite{oscar}, and we introduce a decoupled transformer embedding and decoder step to enable attending over many features per timestep as well as a history of previous embedded states (Figure~\ref{fig:main}).
\modelac\ is the first to bring object-centric navigation to bear on language-guided, manipulation and navigation-based task completion.
We find that \modelac's object-centric navigation and ability to attend across a long time horizon both contribute to its competitive performance with state-of-the-art ALFRED models (Table~\ref{tab:full_ablations}).

Moving forward, we will apply \modelac\ to other benchmarks involving multimodal input through time, such as vision and audio data~\cite{chen20soundspaces}, as well as wider arrays of tasks to accomplish~\cite{virtualhome}.
To further improve performance on the ALFRED benchmark, we could conceivably continue training the \maskrcnn\ model from MOCA~\cite{moca} \emph{forever} by randomizing scenes in AI2THOR~\cite{ai2thor} and having the agent view the scene from randomized vantage points with gold-standard segmentation masks available from the simulator.
For language supervision, we could train and apply a \emph{speaker} model for ALFRED to generate additional training data for new expert demonstrations, providing an initial multimodal alignment for \modelac, a strategy shown effective in VLN tasks~\cite{speaker-follower}.

%% file: 07_limitations.tex
We evaluated \modelac\ only on ALFRED, whose language directives are provided as a one-sided ``recipe'' accomplishing a task.
The \modelac\ architecture is applicable to single-instruction tasks like VLN, as long as auxiliary navigation object targets can be derived from the data as we have done here for ALFRED, by treating the ``recipe'' of step-by-step instructions as empty.
In future work, we would like to incorporate our model on navigation tasks involving dialogue~\cite{cvdn,devries:arxiv18} and real robot platforms~\cite{robotslang} where lifelong learning is possible~\cite{thomason:ijcai15,amazingrace}.
Low-level physical robot control is more difficult than the abstract locomotion used in ALFRED, and poses a separate set of challenges~\cite{blukis:corl19,vln-sim2real}.
By operating only in simulation, our model also misses the full range of \emph{experience} that can ground language in the world~\cite{egl}, such as haptic feedback during object manipulation~\cite{thomason:jair20,thomason:ijcai16,sinapov:icra14}, and audio~\cite{chen20soundspaces} and speech~\cite{harwath2019,rxr} features of the environment. 
Further, in ALFRED an agent never encounters novel object classes at inference time, which represent an additional challenge for successful task completion~\cite{suglia2020imagining}.

The ALFRED benchmark, and consequently the \modelac\ model, only evaluates and considers written English.
\modelac\ inherently excludes people who cannot use typed communication.
By training and evaluating only on English, we can only speculate whether the object-centric navigation methods introduced for \modelac\ will generalize to other languages.
We are cautiously optimistic that, with the success of massively multi-lingual language models~\cite{garrette:acl19}, \modelac\ would be able to train with non-English language data.
At the same time, we acknowledge the possibility of pernicious, inscrutable priors and behavior~\cite{parrots} and the possibility for targeted, language prompt-based attacks~\cite{song:naacl21} in such large-scale networks.

%% file: 99_appendix.tex
\subsection{Additional Auxiliary Losses}\label{appendix:aux_losses}
In this section we describe alternative auxiliary losses that we designed for EmBERT training using ALFRED data. After validation, these configurations did not produce results comparable with the best performing model. This calls for a more detailed analysis of how to adequately design and combine such losses in the complex training regime of the ALFRED benchmark. 
\paragraph{Masked Language Modeling}
The task-oriented language in ALFRED differs from the web crawl text used to train large-scale Transformers. 
We tune our initial model weights using a masked language modeling objective~\cite{bert}.
We mask with a $\%15$ probability a token among the ones in $\alfgoal$ and $\alfinst_t$ at the very last step of a sub-goal.
Differently from captions data or Wikipedia, {\em when} which such supervision should be provided is crucial. 
Given the instruction \textquote{Turn around and walk to the book on the desk}, at the very first timestep of the trajectory it is likely that none of the mentioned objects are visible.
Thus, we assume that at the last step of a subgoal the agent will have in view the objects associated with the instruction. 
We apply the same conditional scaling approach to generate time-dependent language representations $\mathbf{\tilde{L}}$ as the one used in Equation~\ref{eq:conditional_scaling}. 
We denote the masked language modeling loss used for this task by $\mathcal{L}_{MLM}$.

\paragraph{Masked Region Modeling}
This is analogous to the Visual Region Classification (VCR) loss that we integrated in the model. The main difference is that $15\%$ of the visual features are entirely masked (i.e., replaced with zero values) and we ask the model to predict them given the time-dependent representations generated by \modelac\ for them. 

\paragraph{Image-text Matching}
The masked region and language modeling losses encourage the model to learn fine-grained object and language token representations, respectively.
However, we are also interested in global representations that are expressive enough to encode salient information of the visual frames.
For this reason, we design an additional loss $\mathcal{L}_{IM}$.
Given the state representation for the current timestep $t$, \modelac\ predicts whether the current visual features can be associated with the corresponding language features or not. 
We maximize the cosine similarity between the visual features of the current timestep $t$ and the corresponding language features while, at the same time, minimizing the cosine similarity between the current visual features and other language instructions in the same batch. In this task, just like when modeling the robot state, we use $\mathbf{\tilde{L}}_0$ as the language features and $\mathbf{\tilde{L}}_{m+n}$ as the visual features.
We define $\mathcal{L}_{IM}$ the same way as the contrastive loss in CLIP~\cite{clip}. 
However, we expect the model to use the time-dependent representation of the agent state in order to truly understand the meaning of a language instruction. In this case the meaning of an instruction can be appreciated only after several timesteps when the corresponding sequence of actions has been executed. 

\subsection{\modelac\ Asset Licenses}

AI2THOR~\cite{ai2thor} is released under the Apache-2.0 License, while the ALFRED benchmark~\cite{alfred} is released under the MIT License.

